\pdfoutput=1

\documentclass[11pt]{article}

\usepackage{acl}

\usepackage{times}
\usepackage{latexsym}
\usepackage{expex}
\usepackage{amssymb}
\usepackage{amsmath}
\usepackage{amsfonts}
\usepackage{mxedruli}
\usepackage{graphicx}
\usepackage{todonotes}
\usepackage{enumitem}
\usepackage{qtree}
\usepackage{tree-dvips}
\usepackage{multirow}
\setlist{noitemsep}
\usepackage{kotex}
\usepackage[T1]{fontenc}

\usepackage[utf8]{inputenc}
\usepackage{microtype}

\usepackage[russian, english]{babel}
\newcommand{\paren}[1]{\left(#1\right)}
\newcommand{\fe}[2]{feat_#1^#2}
\newcommand{\fo}[2]{form_#1^#2}
\newcommand{\rus}[1]{\foreignlanguage{russian}{#1}}
\newcommand{\kat}[1]{\begin{mxedr}#1\end{mxedr}}

\title{Morphological Reinflection with Multiple Arguments:\\An Extended Annotation schema and a Georgian Case Study}

\author{David Guriel, \ Omer Goldman, \  Reut Tsarfaty \\
Bar-Ilan University \\
\texttt{\{davidgu1312,omer.goldman\}@gmail.com,reut.tsarfaty@biu.ac.il}}

\begin{document}
\maketitle
\begin{abstract}
In recent years, a flurry of morphological datasets had emerged, most notably UniMorph, a multi-lingual repository of inflection tables. %
However, the flat structure of the current morphological annotation schema makes the treatment of some languages quirky, if not impossible, specifically in cases of polypersonal agreement, where verbs agree with multiple arguments using true affixes.
In this paper, we propose to address this phenomenon by expanding the UniMorph annotation schema to a hierarchical feature structure that naturally accommodates complex argument marking.
We apply this extended schema to one such language, Georgian, and provide a human-verified, accurate and balanced morphological dataset for Georgian verbs. The dataset has 4 times more tables and 6 times more verb forms compared to the existing UniMorph dataset, covering all possible variants of argument marking, demonstrating the adequacy of our proposed scheme. Experiments with a  standard reinflection model show that generalization is easy when the data is split at the form level, but extremely hard when splitting along lemma lines.
Expanding the other languages in UniMorph  to this schema is expected to improve both  the coverage,  consistency and  interpretability of this benchmark.
\end{abstract}

\section{Introduction}
In recent years, morphological (re)inflection tasks  have gained a lot of attention in NLP.\footnote{Cf.\ the series of SIGMORPHON shared tasks: \small{\url{https://sigmorphon.github.io/sharedtasks/} }}  Subsequently, several multi-lingual morphological datasets have emerged to allow for the supervised training of morphological models, most notably UniMorph \citep{Unimorph}, that organizes words into inflectional tables,  annotating each inflected word-form with its respective feature-set. %

While western languages are %
widely represented in UniMorph,
many \emph{morphologically rich languages} \citep{tsarfaty-2010,tsarfaty-etal-2020-spmrl} exhibit rich and diverse inflection patterns that make them less compatible with the flat feature-sets in the UniMorph schema.  Concretely, in some cases it is completely impossible to annotate parts of the inflectional paradigm with a flat bundle, as is the case with {\em case stacking}, and in other cases, such as {\em polypersonal agreement}, the annotation solutions provided are  unnatural, non-transparent, and are barely used in practice. 
As a result, languages exhibiting such phenomena are under-represented in UniMorph, and when they are, the inflection tables for these languages are often incomplete.

In this paper we propose a general solution for annotating such structures, thus extending the UniMorph annotation schema to  fully cover a  wider range of morphologically-complex argument-marking phenomena. Following \citet{anderson-1992-morphous}, we propose a so-called {\em layered} annotation of features, where the inflectional features take the form of a {\em hierarchical} structure, in the spirit of formal linguistic frameworks as that of  \citet{johnson,pollard-sag-1994-head,shieber,bresnan-etal-2015-lexical}.
We organize the features of multiple arguments in a hierarchical structure, rather than the current flat structure that accommodates only subject concords. This schema  shift  allows for an adequate annotation  of {\em polypersonal agreement} and of  {\em possessed nominals}, where a word has multiple number and gender features, as well as forms with {\em case stacking}, where a word has multiple cases.
 
We apply the suggested solution to Georgian,
an agglutinative language with a convoluted verbal system, that indicates both subjects and objects with true affixes (rather than clitics that are omittable from the inflection tables). We create a new human-verified dataset for Georgian,  that covers most of the grammatical phenomena in Georgian verbs, %
and includes 118 lemmas, adding up to about 21$k$ verb forms, compared with the 47 lemmas and 3.3$k$ verb forms, some of which are erroneous, currently available in the Georgian UniMorph.

We use the new dataset to train a standard morphological reinflection model %
\citep{silfverberg-hulden-2018-encoder}
and show that
training on the  Georgian inflections currently available in UniMorph is not sufficient for generalizing to the more inclusive set of inflections that are allowed by the new scheme. We conclude that our annotation approach provides a more complete representation of linguistic behaviors, and that our proposed Georgian dataset provides a much better depiction of the morphological phenomena that exist in the data and the computational challenge reflected therein.

We therefore call to apply layered annotation to all currently existing morphological data in UniMorph, to more consistently and transparently capture the linguistic reality and morphological complexity reflected in the worlds languages. %

\section{The Problem: Multiple Arguments}
\label{sec:Background}

\begin{table}[t]
\centering
\small{
\begin{tabular}{llllll}
\multicolumn{6}{l}{\kat{gagi+svebt} (\textit{gagi\v{s}vebt})} \\
\hline
\kat{ga-} & \kat{g-} & \kat{i-} & \kat{+sv} & \kat{-eb} & \kat{-t}  \\
ga- & g- & i- & \v{s}v & -eb & t \\
\textsc{Fut} &\textsc{o2Sg} & \textsc{Trans} & \textsc{let go} & \textsc{Theme} & \textsc{s1Pl} \\
\hline
\multicolumn{6}{l}{\emph{We will let you(sg.) go}}
\end{tabular}
}
\caption{A typical Georgian verb. Note the 2 argument markers,  object (tagged with \textsc{o}) and subject (\textsc{s}).}
\label{tab:verb_example}
\end{table}

Models of morphological reinfection are trained to generate forms within a lemma $L$, given another form and the features of   source\(_i\) and target\(_j\) forms:
\[\paren{\langle \fe{i}{L}, \fo{i}{L}\rangle , \langle \fe{j}{L}, \_\_\_ \rangle } \mapsto \fo{j}{L}\]

\noindent For example, for the Russian lemma \rus{ЛЕТЕТЬ}: %
reinflecting from (\textsc{Prs;1;Sg},\rus{лечу}) to (\textsc{Imp;2;Sg},\rus{лети}) will be represented as: 
\[ 
     \big(\langle
     \textnormal{\textsc{Prs;1;Sg},\rus{лечу}}
     \rangle, 
     \langle
     \textsc{Imp;2;Sg}, 
     \_\_\_
     \rangle
     \big) \mapsto \textnormal{\rus{лети}}
\]

Standardly, the data for training morphological models (e.g., \citealp{transformer2, makarov-clematide-2018-neural}) 
is taken from UniMorph \citep{Unimorph}, a multilingual morphological dataset in which words are grouped by lemma into inflection tables, each word is tagged with an unordered set of morphological features. The features list is shared across languages. The inflection tables are meant to be exhaustive, i.e., covering all possible forms of a lemma, regardless of usability.

Although the features were designed to apply cross-lingually, some blind-spots  exist. Most relevant to our work is the assumption that every feature set includes at most one pronominal feature bundle (i.e., person-gender-number).

However, this assumption does not apply to verbs with object concords, as exhibited in Georgian (see Table~\ref{tab:verb_example}), Inuit and many Bantu languages \textit{inter alia}, nor does it apply to possessed nouns that mark the features of both the possessor and the possessee.
Examples (1a)--(1d) illustrate this: %

\begin{itemize}
        \item[(1)] 
\begin{enumerate}[nosep]
    \item[a.] Georgian: \emph{gagi\v{s}vebt} `We will let you go'\\ (\textsc{subj-1pl}, \textsc{obj-2sg})
    \item[b.]  Turkish: \emph{kedisisin} `you are his cat'\\ (\textsc{Noun-sg}, \textsc{subj-2sg},   %
    \textsc{poss-3sg})
    \item[c.]  Swahili: \emph{ninakupenda} `I love you'\\ (\textsc{subj-1sg}, \textsc{obj-2sg})
    \item[d.]  Hebrew:  \emph{emdata} `her position' \\ (\textsc{Noun-sg}, \textsc{poss-3sg-fem})
\end{enumerate}
\end{itemize}

The solution proposed in UniMorph to annotating these phenomena is via  concatenating several properties into a single string, lacking any internal structure;
e.g., \textsc{Argac2s} indicates  a form with a 2nd person singular accusative argument \citep{sylak-2016-composition}. However, there are at least two  shortcomings to this solution. First, it is not sufficiently transparent.  \textsc{Argac2s} is an opaque string, that does not decompose into the known features licensed by the UniMorph features list (i.e., \textsc{acc}, \textsc{2}, \textsc{sg}). Secondly, and possibly due to this lack of transparency, this annotation hack is hardly ever used in practice. Hence, from all examples in (1), only the Hebrew form is included in UniMorph, and tagged as \textsc{n;sg;fem;pss3s} with multiple possessor features merged into the flat  string \textsc{Pss3s}.%

The crux of the matter is that in the current annotation schema,  complex features assigned to  additional arguments are treated as a single non-decomposable feature, that lack any internal structure, unlike the features of the main (so-called `internal') argument, that are individually spelled out. We argue that the lack of transparency and  usability are due to 
the misrepresentation of the inherently {\em hierarchical} and {\em compositional} structure of the features in such forms.
We suggest to explicitly annotate these forms with features that are all explicitly composed of the same primitive features.%

All in all, the lack of a sufficiently expressive annotation standard leads to a data distribution that is skewed, unrealistically simple, and, when language-specific annotation solutions are painfully needed, they suffer from inconsistencies and ad-hoc decisions. For these reasons, we set out to extend the UniMorph annotation schema  to accommodate all such cases and to enable a proper coverage of languages, such as Georgian and many others.

\section{The Proposed Schema}

\begin{table*}[t]
    \centering
    \begin{tabular}{lcc}
         \hline
          & \textbf{Flat structure} & \textbf{Hierarchical Structure} \\\hline
         Georgian: \emph{gagi\v{s}vebt} & \multirow{4}{*}{\scalebox{0.7}{ \Tree [ \textsc{fut} \textsc{argno1p} \textsc{argac2s} ].V }} & \multirow{4}{*}{\scalebox{0.7}{ \Tree [ \textsc{fut} [ \textsc{1} \textsc{pl} ].\textsc{nom} [ \textsc{2} \textsc{sg} ].\textsc{acc} ].V }} \\ Trans: `We will let you go' \\
         Args: \textsc{subj-1pl}, \textsc{obj-2sg} \\
         \\
         \hline
         Turkish: \emph{kedisisin} & \multirow{4}{*}{\scalebox{0.7}{ \Tree [ \textsc{sg} \textsc{argno2s} \textsc{pss3s} ].N }} & \multirow{4}{*}{\scalebox{0.7}{ \Tree [ \textsc{sg} [ \textsc{2} \textsc{sg} ].\textsc{nom} [ \textsc{3} \textsc{sg} ].\textsc{poss} ].N }} \\ Trans: `you are his cat' \\
         Args: \textsc{Noun-sg}, \textsc{subj-2sg},
        \textsc{poss-3sg} %
        \\
        \\\hline
         Swahili: \emph{ninakupenda} & \multirow{4}{*}{\scalebox{0.7}{ \Tree [ \textsc{prs} \textsc{argno1s} \textsc{argac2s} ].V }} & \multirow{4}{*}{\scalebox{0.7}{ \Tree [ \textsc{prs} [ \textsc{1} \textsc{sg} ].\textsc{nom} [ \textsc{2} \textsc{sg} ].\textsc{acc} ].V }} \\ Trans: `I love you' \\
         Args: \textsc{subj-1sg}, \textsc{obj-2sg} %
         \\
         \\\hline
         Hebrew: \emph{emdata} & \multirow{4}{*}{\scalebox{0.7}{ \Tree [ \textsc{sg} \textsc{pss3s} \textsc{fem} ].N }} & \multirow{4}{*}{\scalebox{0.7}{ \Tree [ \textsc{sg} [ \textsc{3} \textsc{sg} \textsc{fem} ].\textsc{poss} ].N }} \\ Trans: `her position' \\
         Args: \textsc{Noun-sg}, \textsc{poss-3sg-fem} %
         \\
         \\\hline
    \end{tabular}
    \caption{Examples for word-forms with multiple argument agreements. On the left we present the flat structure  currently employed in UniMorph. All examples save Hebrew  are  not included in the  UniMorph inflection tables, presumably due to their lack of transparency. On the right we present our proposed hierarchical structure, which is more transparent, and also ammenable for compositional generalization.}
    \label{tab:examples}
\end{table*}

We propose to extend the UniMorph annotation schema to cover multiple pronominal feature-bundles in the same word-form, via a {\em layering} approach, originally proposed for morphological systems by \citet{anderson-1992-morphous}.
\citeauthor{anderson-1992-morphous}  suggests to arrange the {\em morphosyntactic representation} (MSR) of words in a hierarchy (dubbed {\em layers}) of features, in the sense that every element of the unordered set of features can be composed of another unordered set of  features.  %
That is, a general feature annotation  looks as in (2a). A specific transitive  verb annotation could be as depicted  in (2b):

\begin{itemize}
    \item[(2)] 
\begin{itemize}
    \item[a.] $[f_1, f_2,..., [F_i: f_{i_1}, f_{i_2},... [F_j: f_{j_1}.. ] ] ]$
    \item[b.] $[V, Tense,$ \\ \hphantom{~~~} $[nom: Per, Num, Gen],$ \\ \hphantom{~~~} $[acc: Per, Num, Gen]]$
\end{itemize}
\end{itemize}
{This hierarchical feature structure is reminiscent of {\em unification grammars} or {\em attribute-value grammars} \cite{shieber,johnson} that are extensively used in  syntactic theories such as GPSG, HPSG, and resemble the f-structures in LFG \cite{gpsg,pollard-sag-1994-head,bresnan-etal-2015-lexical}.}

Here we employ these structures to organize the features of morphologically-marked arguments   hierarchically, so an argument is characterized by a feature composite of all features pertaining to that argument. 
That is, each argument's feature-bundle os specifically marked with the argument it belongs to, and is decomposed into the primitive features licensed by the UniMorph scheme. It also  homogeneously annotates  the different kinds of arguments, in contrast with the current schema where the subject features are assigned to the verb directly. 
Thus, the English form \textit{thinks} previously annotated as \textsc{v;prs;3;sg} will be annotated as \textsc{v;prs;nom(3;sg)}. In languages that mark multiple arguments, different kinds of arguments can be marked with their  feature-bundles without conflicts.
The proposed schema thus facilitates the annotation of the poorly-treated or untreated phenomena as illustrated in  (1). These are, respectively:
\begin{itemize}
        \item[(3)] 
\begin{enumerate}[nosep]
    \item[a.]  Georgian: \emph{gagi\v{s}vebt} `We will let you go'\\  \textsc{v;fut;nom(1;pl);acc(2;sg)}
    \item[b.] Turkish: \emph{kedisisin} `you are his cat'\\ \textsc{n;sg;nom(2;sg);poss(3;sg)}
    \item[c.]  Swahili: \emph{ninakupenda} `I love you'\\ \textsc{v;prs;nom(1;sg);acc(2;sg)}
    \item[d.]  Hebrew: \emph{emdata} `her position'\\ \textsc{n;sg;poss(3;sg;fem)}
\end{enumerate}
\end{itemize}

Table~\ref{tab:examples} compares the annotation of these examples in the current UniMorph schema compared with our proposed annotation schema.\footnote{Although not explicitly shown here, annotation of case stacking is also possible with our approach, while non-hierarchical annotations do not account for such cases. For example, Korean 교사에게이 can be tagged as \textsc{n;sg;nom(dat)}.} The hierarchical structures, beyond being more transparent, opens the door further for future study on compositional generalization in morphology.

The resemblance of our proposed schema to ideas  in other fields of theoretical linguistics, most prominently to the \textit{f-structure} in LFG \citep{bresnan-etal-2015-lexical} and to the nested \textit{Attribute-Value matrices} in HPSG \citep{pollard-sag-1994-head}, points to a natural interface with further syntactic and semantic annotations downstream.

\section{A Case Study from Georgian}
\paragraph{Linguistic Background}
Georgian is an agglutinative language with a verbal system that makes a vast use of affixes to convey a wide array of meanings, both inflectional and derivational %
(see Table~\ref{tab:verb_example}).
The Georgian verbal paradigm is divided into 5 classes known as: transitive, intransitive, medial, indirect and stative \cite{georgianbook}. 
The verbs are inflected to reflect 12 Tense-Aspect-Mood (TAM) combinations (traditionally known as \textit{screeves}) sorted into 4 series: present and future, aorist, perfective, and the imperative. Each series has its own morpho-syntactic characteristics, most notably  split-ergativity is manifested in the aorist. 

The characteristic most essential to this work is that Georgian verbs always agree on person and number with the direct and indirect objects, on top of the subject-verb agreement. 
The Georgian data in UniMorph follows the  convention of including  objects  {\em only} in third person singular %
--- thus failing to provide a comprehensive coverage of the word-forms that can be attested in the language.

Additional issues with the current morphological data in UniMorph for Georgian verbs are: sparsity, as it includes only 47 inflection tables; lack of diversity, as all table are  from the transitive class; and lack of accuracy, as the data was produced automatically without verification by native speakers.

\paragraph{Data Annotation}
A key contribution of this work is the creation of a new dataset for Georgian that follows the layered annotation schema and addresses the other  shortcomings just described. 
We selected a list of 118 verb lemmata from all different classes.\footnote{We based the list of verbs on those whose inflection tables appear on \citet{georgianbook} and added some commonly-used verbs suggested by native speakers.}
Every verb was manually annotated with its stem, its thematic affix and  principal parts,   
to automatically generate the full  inflection tables.

 This automatic generation of Georgian verbs is prone to some errors, for instance, in accounting for idiosyncratic phonologically-conditioned stem changes. Hence, we ran our data through 3 native Georgian speakers to assert its correctness, or fix when needed. In cases where speakers were unsure we used a Georgian morphological analyzer \cite{glc2} for consultation. %
 In cases of  disagreement, we used a majority vote among the speakers.
On average, at least one speaker was uncertain in about 5\% of the forms, but a disagreement that necessitated a majority vote occurred only on about 0.7\% of the cases.

Table~\ref{tab:datastat} summarizes the statistics over our annotated data. In total, we produced 21,054 verb forms, of 118 lemmata. The data is quite evenly balanced across the classes, with more verbs drawn from the more frequent transitive class. For comparison, the current UniMorph data has fewer lemmas, 3,300 forms, and includes only verbs that are transitive.\footnote{All our data is publicly available at \url{https://github.com/Onlp/GeorgianMorphology}.}

\begin{table}[t]
\centering
\resizebox{\columnwidth}{!}{
\begin{tabular}{lccccc}
\hline
 & \textbf{Trans.} & \textbf{Intrans.} & \textbf{Med.} &\textbf{Indi.} & \textbf{Stat.}\\
\hline
\#Infl. Tables & 40 & 21 & 29 & 16 & 12 \\
\#Verb Forms & 12506 & 2560 & 3132 & 2626 & 230 \\
\hline
\end{tabular}
}
\caption{Distribution of the  Georgian verbs over  classes.}
\label{tab:datastat}
\end{table}

\section{Experiments}
To assess the usability of our dataset, %
we trained a standard reinflection model, the
character-level LSTM of %
\citet{silfverberg-hulden-2018-encoder}, on our data.\footnote{For hyper-parameters tuning see Appendix~\ref{sec:hypers}.}
We {sampled} from our data 2 datasets  for training  morphological reinflection models, 
containing train, validation and test sets in sizes  8$k$, 1$k$ and 1$k$ examples, respectively. 
Following \citet{goldman2021unsolving}, one dataset employed  an easier form-split, i.e., no forms appear in both train and test,\footnote{This is the splitting method used in SIGMORPHON's shared tasks on reinflection (e.g., \citealp{cotterell-etal-2018-conll}).}  and the other with the more challenging lemma-split, where lemmas from  train, dev and test are disjoint.
To assess the generalization capacity we varied the sources of both the train and test sets.\footnote{We harmonized the train and test features vocabulary, so that the old data bears the new scheme. So the only difference between Original and New is in which forms are  included.}
We report 2 evaluation metrics: {\em accuracy} over exact matches, and {\em average edit distance} from gold.

\paragraph{Results and Analysis}
Table \ref{tab:Accuracy} presents the model's performance for all train-test combinations. It shows that the model's performance on the new data (top line  combination) is largely on par comparing to its performance over training and testing on UniMorph's original data (bottom combination). However, the model generalizes poorly from the original partial data to the forms in our test set which reflect the entire Georgian inflectional system.  Generalization from our data to UniMorph's set is a lot better. %
The results also show that the splitting method is crucial for success of the model, as it inflects easily to unseen forms, but much harder when inflecting forms in a previously unseen lemma.\footnote{For learning curves on the splits see Appendix~\ref{sec:curves}.} These results corroborate the results of \citet{goldman2021unsolving} regarding the difficulty of lemma-split data.
Although the accuracy over the lemma split data is negligible, the average edit distance in that case points again to the conclusion that generalization from UniMorph to our data is harder that the other way around.

\begin{table}[t]
\centering
\resizebox{\columnwidth}{!}{
\begin{tabular}{ll|cc|cc}
\hline
\multirow{2}{*}{\textbf{Train Set}} & \multirow{2}{*}{\textbf{Test set}} & \multicolumn{2}{c}{\textbf{Form Split}} & \multicolumn{2}{c}{\textbf{Lemma Split}}\\
& & \textbf{Acc} & \textbf{Avg ED} & \textbf{Acc} & \textbf{Avg ED} \\
\hline
New & New & 94.9\% & 0.15 & 1.3\% & 4.66 \\
New & Original & 84.7\% & 0.3 & 0.3\% & 4.39 \\
Original & New & 35.2\% & 1.36 & 0.0\% & 6.22 \\
Original & Original & 99.3\% & 0.01 & 0.0\% & 6.13 \\
\hline
\end{tabular}
}
\caption{Accuracy ({\bf Acc}, higher is better) and Average Edit Distance ({\bf Avg ED}, lower is better) for morphological reinflection on different train-test combinations.}
\label{tab:Accuracy}
\end{table}

\paragraph{Error Analysis} To provide insights into the challenge of reinflecting  morphologically complex forms, we manually sampled the erroneous output of the model trained and tested over our lemma-split data, to draw insights on the points of failure. In many cases the model succeeded in copying and modifying the verb stem, but failed to output the other morphemes correctly. Sometimes the errors were due to inflection to an incorrect TAM combination of the same lexeme, and sometimes the inflection was done to the correct TAM but to a different derivationally-related lemma (e.g. change of voice in addition to the change of TAM). We conclude that the fact that our datasets include lemmas from diverse classes that may have derivational relations makes the inflection task significantly harder. Interestingly, the model managed to predict the correct subject and object affixes most of the time.

\section{Conclusion}

This paper
proposes a transition of the UniMorph annotation standard to a layered hierarchical annotation of features. This revised schema caters for complex marking phenomena %
including multiple pronominal agreement. %
We apply it to Georgian,
and construct a corresponding new dataset %
that is large, balanced,  complete with respect to grammatical phenomena in the Georgian verb system and verified by native-speakers. 
Our experiments with a standard reinflection model on the old and new Georgian datasets shows that 
the old UniMorph dataset does not  generalize well to  the new test-set, due to its partial coverage.
This work is intended to encourage the community to extend the annotation of different languages to include phenomena such as polypersonal agreement and others that can be dealt with using a hierarchical annotation, %
ultimately leading to more complete and consistent benchmarks for studying  non-trivial and less-explored areas of computational morphology.

\section*{Acknowledgements}

The first author would like to thank the native Georgian speakers: Simon Guriel, Silvia Guriel-Agiashvili and Nona Atanelov for their invaluable help in the %
data annotation process. 
This research was funded by the European Research Council under the European Union’s Horizon 2020 research and innovation programme (grant agreement No.\ 677352)  and by a research grant from the Ministry of Science and Technology (MOST) of the Israeli Government, for which we are grateful.

\bibliography{anthology,custom}
\bibliographystyle{acl_natbib}
\nocite{glc1,glc2}

\newpage
\appendix

\section{Learning Curves}
\label{sec:curves}

\begin{figure}[ht]
\centering
	\includegraphics[width=0.99\columnwidth]{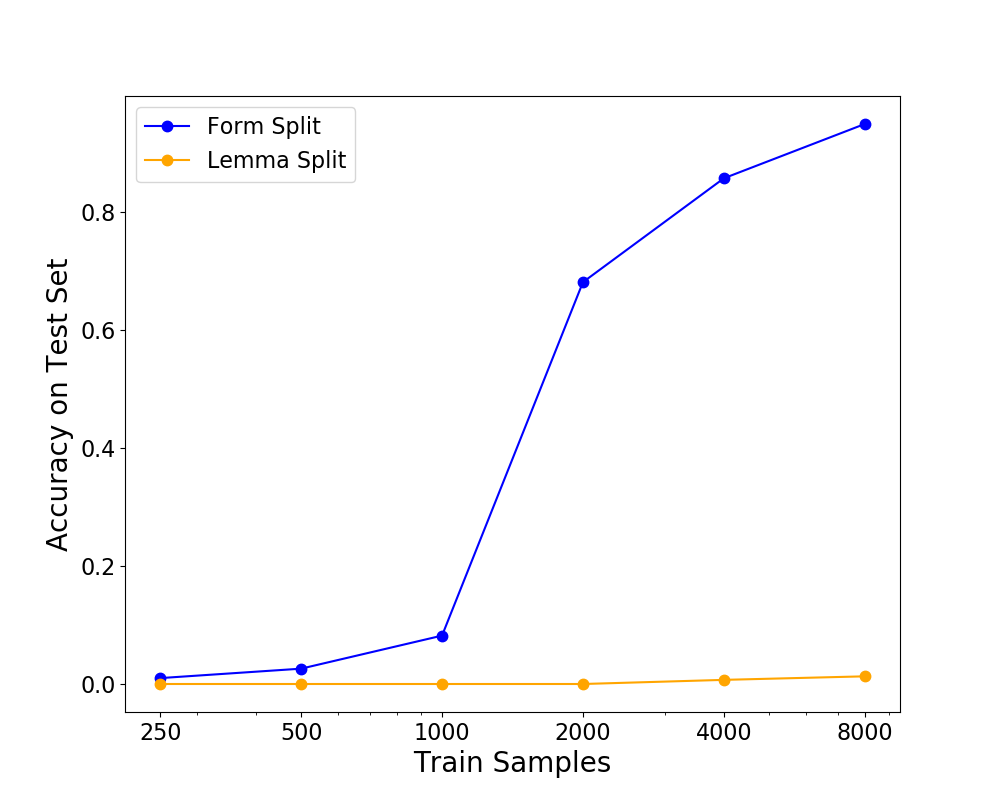}
	\caption{Inflection accuracy over \textit{form-split} and \textit{lemma-split} test sets as a function of train set size.}
	\label{fig:lc}
\end{figure}

Fig.\ \ref{fig:lc} exemplifies the sufficiency of our dataset for training an inflection model on form-split data as doubling the data amount from 4,000 to 8,000 yields relatively minor improvement. It also shows that for the lemma-split data, the model completely fails. It starts improve marginally with more than 2,000 examples,
although its performance remains far from satisfactory. This leaves room for exploration of bootstrapping and augmentation methods or more sophisticated modeling to improve results.

\section{Tech-Spec}

All algorithms described in the paper were executed on a single machine equipped with one NVIDIA TITAN Xp GPU, 16 Intel i7-6900K(3.20GHz) CPUs and 126GB RAM. Since the LSTM algorithm was implemented on DyNet, there was no need of the GPU, and all the calculations were done using only the CPU.

\section{Hyper Parameters}
\label{sec:hypers}

\begin{enumerate}
    \item Embedding size = 100
    \item Hidden state size = 100
    \item Attention size = 100
    \item Number of LSTM layers = 1
\end{enumerate}
During training, we experimented with several values for the hyper-parameters detailed above. However, for all the combinations we tried, the results barely changed both at the form-split setting and the lemma-split setting.

\end{document}